\documentclass[10pt,twocolumn,letterpaper]{article}

\usepackage{cvpr}              % To produce the CAMERA-READY version

\usepackage[utf8]{inputenc} % allow utf-8 input
\usepackage[T1]{fontenc}    % use 8-bit T1 fonts
\usepackage{graphicx}
\usepackage{url}            % simple URL typesetting
\usepackage{booktabs}       % professional-quality tables
\usepackage{amsfonts}       % blackboard math symbols
\usepackage{nicefrac}       % compact symbols for 1/2, etc.
\usepackage{microtype}      % microtypography
\usepackage{amsmath}
\usepackage{enumitem}
\usepackage[dvipsnames,table,xcdraw]{xcolor}
\usepackage{pifont} 
\usepackage{multirow}
\usepackage{amssymb}
\usepackage{caption}
\usepackage{subcaption}
\usepackage{wrapfig}  % 在导言区导入包
\usepackage{lipsum}   % 用于示例文字

\usepackage{placeins}
\usepackage{utfsym}
\usepackage{fontawesome5} % for \faImage etc.

\definecolor{myblue}{HTML}{D6E4F4}  % soft blue
\definecolor{background}{HTML}{FFFFFF} % E0FFFF

\definecolor{fusion}{HTML}{FFA500} % E0FFFF‘’‘’‘’‘’‘’‘

\definecolor{textcolor}{HTML}{3E89B5}     % blue
\definecolor{imagecolor}{HTML}{4FB596}     % green
\newcommand{\textsymb}{\textcolor{textcolor}{\faIcon*[regular]{file}}} %{\ding{46}}}  
\newcommand{\imagesymb}{{\textcolor{imagecolor}{\faImage}}}
\newcommand{\promptfuse}{\textcolor{fusion}{\scalebox{0.8}{\ding{71}}}}
\newcommand{\promptfuseS}{\textcolor{fusion}{\raisebox{-0.3ex}{\scalebox{0.8}{\ding{71}}\textsubscript{\textbf{s}}}}}
\newcommand{\latentfuse}{\textcolor{fusion}{\scalebox{0.8}{\ding{70}}}}

\definecolor{softgreen}{HTML}{4CAF50}  % muted professional green
\definecolor{softred}{HTML}{E57373}    % gentle desaturated red
\newcommand{\yesmark}{{\textcolor{softgreen}{\ding{51}}}} % ✓
\newcommand{\nomark}{{\textcolor{softred}{\ding{55}}}}   % ✗

\definecolor{cvprblue}{rgb}{0.21,0.49,0.74}
\usepackage[pagebackref,breaklinks,colorlinks,allcolors=cvprblue]{hyperref}

\def\paperID{} % *** Enter the Paper ID here
\def\confName{CVPR}
\def\confYear{2026}
              
\title{GraphVLM: Benchmarking Vision Language Models for  Multimodal Graph Learning 
}

\author{
Jiajin Liu$^{1,2}$,
Dongzhe Fan$^{1,2}$,
Chuanhao Ji$^{1,4}$,
Daochen Zha$^{3}$,
Qiaoyu Tan$^{1}$\footnotemark[1] \\
$^{1}$NYU Shanghai,
$^{2}$New York University, 
$^{3}$Rice University,
$^{4}$East China Normal University \\
\texttt{\{jiajinliu, qiaoyu.tan\}@nyu.edu}
}

\begin{document}
\maketitle

\renewcommand{\thefootnote}{\fnsymbol{footnote}}
\footnotetext[1]{Corresponding author}

\begin{abstract}

Vision-Language Models (VLMs) have demonstrated remarkable capabilities in aligning and understanding multimodal signals, yet their potential to reason over structured data, where multimodal entities are connected through explicit relational graphs, remains largely underexplored. Unlocking this capability is crucial for real-world applications such as social networks, recommendation systems, and scientific discovery, where multimodal information is inherently structured.
To bridge this gap, we present GraphVLM, a systematic benchmark designed to evaluate and harness the capabilities of VLMs for multimodal graph learning (MMGL). GraphVLM investigates three complementary paradigms for integrating VLMs with graph reasoning: (1) VLM-as-Encoder, which enriches graph neural networks through multimodal feature fusion; (2) VLM-as-Aligner, which bridges modalities in latent or linguistic space to facilitate LLM-based structured reasoning; and (3) VLM-as-Predictor, which directly employs VLMs as multimodal backbones for graph learning tasks.
Extensive experiments across six datasets from diverse domains demonstrate that VLMs enhance multimodal graph learning via all three roles. Among these paradigms, VLM-as-Predictor achieves the most substantial and consistent performance gains, revealing the untapped potential of vision–language models as a new foundation for multimodal graph learning. The benchmark code is publicly available at \url{https://github.com/oamyjin/GraphVLM}.

\end{abstract}    
\section{Introduction}
\label{sec:intro}

In recent years, Vision-Language Models (VLMs) have made impressive strides in learning and interpreting information across multiple modalities. Some models~\cite{clip, blip, li2023blip, girdhar2023imagebind} are pre-trained for multimodal alignment, such as CLIP~\cite{clip}, and excel at fusing image and text information for representation learning. Some other models~\cite{bai2023qwenvlversatilevisionlanguagemodel, liu2023llava, openai2024gpt4technicalreport, geminiteam2025geminifamilyhighlycapable}, such as Qwen-VL~\cite{bai2023qwenvlversatilevisionlanguagemodel}, are designed to process multimodal inputs, combining both text and visual information to enhance understanding and reasoning capabilities. Despite their success, most VLMs focus on pairwise modality alignment, typically between image and text, without modeling the relational structure that connects entities in real-world data. However, such structural relationships are fundamental in many domains, including social networks~\cite{Wu_2021, sharma2024survey}, recommendation systems~\cite{liu2025pone, anand2025survey}, and scientific knowledge graphs~\cite{yasunaga2021qa, su2022gnn}, where understanding inter-node dependencies is essential for reasoning beyond isolated pairs.

\begin{figure*}[t]
\centering
\includegraphics[width=1\textwidth]{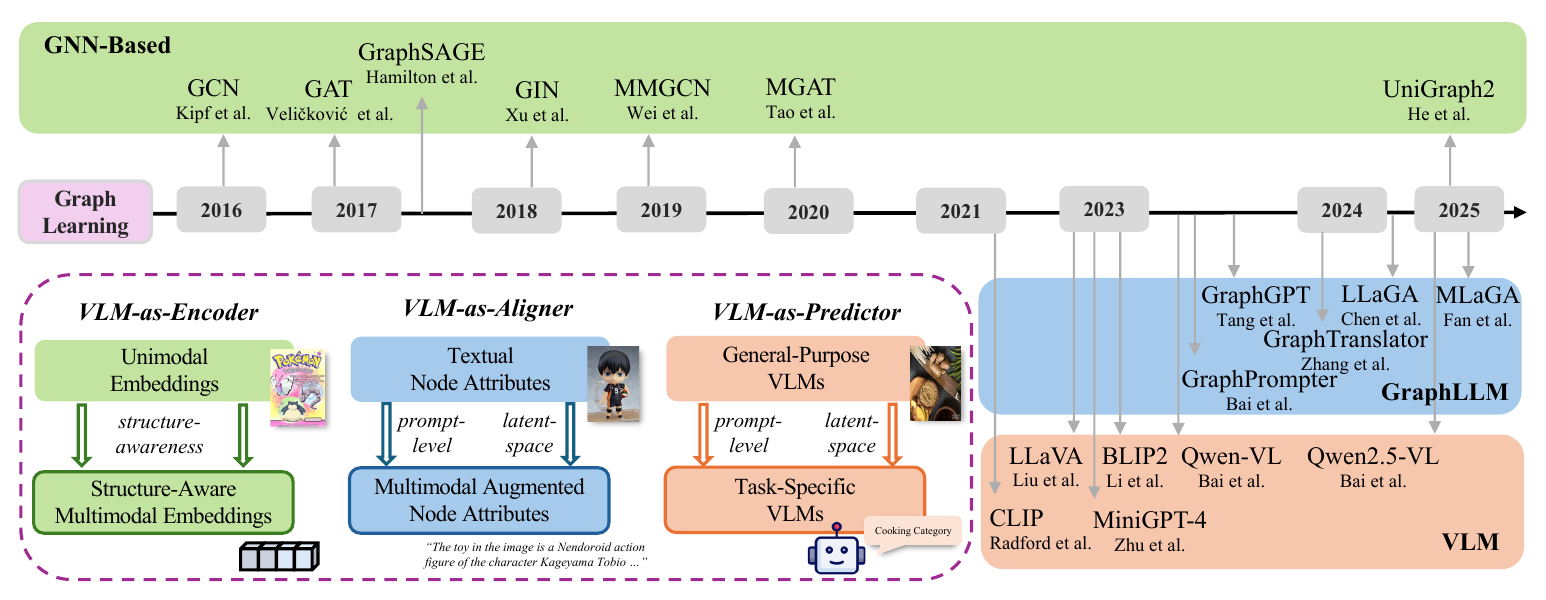}
\caption{Overview of the GraphVLM benchmark with a timeline of graph learning research. Existing graph learning methods are categorized into three groups based on the prediction backbone. The bottom-left corner illustrates the functional roles VLMs play in each category.}
\label{fig:graphVLM}
\end{figure*}

To capture such relationships, Multimodal graph learning (MMGL) has emerged as a promising paradigm that combines heterogeneous node attributes with relational structures, enabling richer representations for prediction and recommendation~\cite{zhu2024multimodalgraphbenchmark, yan2024graph}. Existing MMGL efforts largely fall into two families: \textit{(1) GNN-based models}~\cite{kipf2016gcn, hamilton2017graphSAGE, velivckovic2017gat, uniGraph2} form the traditional paradigm, typically incorporating multimodal node features extracted from pre-trained VLMs (e.g., CLIP) to enhance message passing; \textit{(2) LLM-based models}~\cite{chen2024llaga, tang2024graphgpt, mlaga2025, sun2025graphiclunlockinggraphlearning, fang2024uniglmtrainingunifiedlanguage, fang2024gaugllmimprovinggraphcontrastive, liu2026graphsearchagenticsearchaugmentedreasoning} reason over text-attributed graphs and can be extended to multimodal settings by converting images into textual or latent signals that LLMs can process. 
%originally developed for text-attributed graphs but can be extended to multimodal settings by converting visual signals into textual or latent representations that large language models can reason over.

Surprisingly, despite VLMs' inherent strengths in multimodal alignment and cross-modal understanding, \textbf{the third paradigm--\textit{VLM-based models}}, directly employing VLMs as reasoning backbones on multimodal graphs--remains largely underexplored. This gap motivates a core research question: \textit{How and when can VLMs be effectively leveraged to perform multimodal graph learning?}
%which directly employ vision–language models as reasoning backbones. 
%Given the intrinsic multimodal reasoning ability of VLMs, this gap raises a fundamental question: \textbf{\textit{How and when can VLMs be effectively leveraged to tackle multimodal graph learning tasks?}}

Answering this question is non-trivial due to two major hurdles, summarized in Table~\ref{tab:benchmark_comparison}.
%, investigating this research question faces two two major hurdles, as summarized in Table~\ref{tab:benchmark_comparison}.
\ding{182}~\textit{Fragmented baselines and shallow multimodal fusion.} Existing MMGL literature lacks a unified evaluation pipeline, making it difficult to compare GNN-, LLM-, and VLM-based models fairly. Many recent methods, such as UniGraph2~\cite{uniGraph2} and MLaGA~\cite{mlaga2025}, have not been jointly benchmarked, obscuring insights into modality–structure interactions. Furthermore, most GNN-based approaches rely on naive feature concatenation, leaving open questions about how multimodal features should be fused with graph topology.
%Although LLM-based approaches have recently emerged as a promising direction for graph learning, they have not been systematically evaluated under a unified benchmark, leaving their effectiveness for MMGL unclear. Moreover, recent state-of-the-art multimodal graph learning methods such as UniGraph2~\cite{uniGraph2} and MLaGA~\cite{mlaga2025} have yet to be evaluated under common benchmarks, further hindering a holistic understanding of MMGL paradigms. Even for GNN-based methods, which are more commonly benchmarked, modality fusion is often restricted to naive embedding-level concatenation—providing little insight into how multimodal features interact with structural information.
\ding{183}~\textit{Under-explored potential of VLMs in structural reasoning.} Most existing evaluations restrict VLMs to simple zero-shot inference and do not investigate their capacity to act as trainable backbones for graph learning tasks. Moreover, prior work rarely investigates VLMs as multimodal aligners—that is, models capable of integrating visual and textual signals within graph contexts via latent-space fusion or prompt-level transformation. Together, these limitations significantly underestimate the full potential of VLMs when combined with graph structure. 
%Current evaluations restrict VLMs to simple zero-shot roles without fine-tuning for graph tasks, thereby underestimating the potential of VLM as backbones paradigms. In addition, existing studies seldom examine their ability to serve as aligners that integrate visual and textual modalities within graph contexts, either at the prompt or latent level. 

%Even for GNN-based methods, which are more commonly benchmarked, modality fusion is typically limited to naive embedding-level concatenation, failing to capture how multimodal features interact with structural information. 

%Existing benchmarks primarily focus on GNN-based methods, where modality fusion occurs only through simple embedding-level concatenation, overlooking how multimodal features interact with graph structure. Meanwhile, the emerging and widely adopted LLM-based category, which directly employs VLMs for graph reasoning, has not been comprehensively evaluated under a unified protocol, leaving its true potential unexplored. Furthermore, recent LLM-based graph reasoning methods such as UniGraph2~\cite{uniGraph2} and MLaGA~\cite{mlaga2025} remain unevaluated under common benchmarks. \ding{183}~\textit{Under-explored potential of VLMs} — current evaluations restrict VLMs to simple zero-shot roles without fine-tuning for graph tasks, thereby underestimating the potential of VLM as backbones paradigms. Furthermore, existing studies often overlook their ability to serve as aligners that integrate visual and textual modalities within graph contexts, either at the prompt or latent level. 

To address these gaps, we introduce \textbf{GraphVLM}, a comprehensive benchmark that systematically evaluates multimodal graph learning through the lens of \textit{VLM functional roles}. GraphVLM provides a unified protocol spanning GNN-, LLM-, and VLM-based approaches under consistent dataset settings and evaluation criteria. 
%unifies and compares three major paradigms, where VLMs act as \textit{encoders}, \textit{aligners}, and \textit{predictors}, providing a consistent evaluation protocol across \textit{GNN-based}, \textit{LLM-based}, and \textit{VLM-based} approaches. 
Specifically, GraphVLM explores three complementary paradigms: (1) \textbf{\textit{VLM-as-Encoder}}, which employs a pre-trained vision–language model (PVLM) to encode multimodal node features and further fine-tunes it under graph learning paradigms to investigate how structure–modality alignment enhances GNN-based representations; (2) \textbf{\textit{VLM-as-Aligner}}, which examines both latent-space alignment (injecting PVLM multimodal embeddings directly into model input space) and prompt-level alignment (transforming visual information into textual form for LLM reasoning);
%two complementary strategies: latent-space alignment, where multimodal embeddings by PVLMs are directly injected into the model’s input space, and prompt-level alignment, where visual information is transformed by VLMs into textual form before integration into the prompt; 
and (3) \textbf{\textit{VLM-as-Predictor}}, which directly fine-tunes VLMs as task-specific backbones, augmented with structural signals via prompts or latent-space fusion to enable structure-aware multimodal learning on graphs. 
%applies supervised fine-tuning (SFT) and integrates structural cues both explicitly at the prompt level and implicitly within the latent space, enabling VLMs to reason more effectively over multimodal graph structures.

Extensive experiments across six multimodal graph datasets and a broad suite of MMGL methods lead to a clear conclusion:
VLM-as-Predictor consistently achieves the best performance, outperforming GNN- and LLM-based models when enriched with graph-aware prompt or latent-space information. This reveals that VLMs, when properly aligned with graph structure, can serve as powerful backbones for multimodal graph learning and open new directions toward VLM-driven graph reasoning frameworks.
%Extensive experiments across six multimodal graph datasets and a broad suite of MMGL methods provide a clear answer. We find that \textbf{VLM-as-Predictor, when VLMs are fine-tuned with structure-aware prompt or latent-level information, consistently outperforms both GNN- and LLM-based methods.} 
%This indicates that VLMs serve as a more suitable backbone for multimodal graph learning, paving the way for future research toward VLM-based graph reasoning frameworks.

Our main contributions are summarized below:

% Specifically, we investigate how VLMs serve as feature encoders by jointly considering modality–structure alignment, and further examine their roles as aligners and predictors under prompt-based and fine-tuned settings, offering deeper insights into their structure-aware multimodal reasoning capabilities.

%\begin{table}[t]
%\centering
%\caption{Comparison of existing multimodal graph benchmarks and their coverage of modalities, backbones, and VLM roles.}
%\resizebox{\linewidth}{!}{
%\begin{tabular}{l|cc|ccc|ccc}
%\toprule
%\multirow{2}{*}{\textbf{Benchmarks}} & \multicolumn{2}{c|}{\textbf{Modality}} & \multicolumn{3}{c|}{\textbf{Backbone}} & \multicolumn{3}{c}{\textbf{VLM Role}} \\
%\cmidrule(lr){2-3} \cmidrule(lr){4-6} \cmidrule(lr){7-9}
% & Text & Vision & GNN & LLM & VLM & Encoder & Aligner & Predictor \\
%\midrule
%GL-Bench~\cite{li2024glbench} & \checkmark &  & \checkmark & \checkmark &  &  &  &  \\
%MM-Bench~\cite{zhu2024multimodalgraphbenchmark} & \checkmark & \checkmark & \checkmark &  &  & \checkmark &  &  \\
%MAGB~\cite{yan2024graph} & \checkmark & \checkmark & \checkmark &  & \checkmark & \checkmark &  & \checkmark \\
%\midrule
%\textbf{GraphVLM} & \checkmark & \checkmark & \checkmark & \checkmark & \checkmark & \checkmark & \checkmark & \checkmark \\
%\bottomrule
%\end{tabular}
%}
%\label{tab:benchmark_comparison}
%\end{table}

\begin{table}[t]
\centering
\resizebox{\linewidth}{!}{
\begin{tabular}{l|ccc|cccccc}
\toprule
\multirow{3}{*}{\textbf{Benchmarks}} 
& \multicolumn{3}{c|}{\textbf{Backbones}} 
& \multicolumn{6}{c}{\textbf{VLM Roles}} \\

\cmidrule(lr){2-4} \cmidrule(lr){5-10}
%& \textbf{GNN} & \textbf{LLM} & \textbf{VLM} 
& \multirow{2}{*}{\textbf{GNN}}
& \multirow{2}{*}{\textbf{LLM}}
& \multirow{2}{*}{\textbf{VLM}}
& \multicolumn{2}{c|}{\textbf{Encoder}} 
& \multicolumn{2}{c|}{\textbf{Aligner}} 
& \multicolumn{2}{c}{\textbf{Predictor}} \\

\cmidrule(lr){5-6} \cmidrule(lr){7-8} \cmidrule(lr){9-10}
& & & 
& Pre-trained & Fine-tuned 
& Prompt & Latent 
& Zero-shot & SFT \\
\midrule
MM-Bench~\cite{zhu2024multimodalgraphbenchmark} 
& \yesmark & \nomark & \nomark 
& \yesmark & \nomark 
& \nomark & \nomark 
& \nomark & \nomark \\

MAGB~\cite{yan2024graph} 
& \yesmark & \nomark & \yesmark 
& \yesmark & \nomark 
& \nomark & \nomark 
& \yesmark & \nomark \\

\midrule
\textbf{GraphVLM} 
& \yesmark & \yesmark & \yesmark 
& \yesmark & \yesmark 
& \yesmark & \yesmark 
& \yesmark & \yesmark \\
\bottomrule
\end{tabular}
}

\caption{Comparison of existing multimodal graph benchmarks and their coverage of MMGL method backbones and VLM roles.}
\label{tab:benchmark_comparison}
\end{table}

\begin{itemize}[noitemsep,leftmargin=*,label=\ding{70}]
\item \textbf{Systematic exploration of VLMs for multimodal graph learning.} 
We present the first systematic benchmark that explores the untapped potential of VLMs for multimodal graph learning. 
The benchmark unifies three VLM functional roles to investigate how model finetuning and multimodal fusion contribute to structure-aware multimodal graph reasoning.

\item \textbf{Unified and comprehensive benchmark suite.} 
We establish a large-scale benchmark covering GNN-, LLM-, and VLM-based approaches on six multimodal graph datasets. 
Our framework considers both \textit{prompt-level} and \textit{latent-space} fusion strategies, enabling fair and consistent comparison across paradigms.

\item \textbf{In-depth multi-dimensional analysis.} 
We conduct extensive analyses across three complementary perspectives—\textit{modality fusion}, \textit{structure integration}, and \textit{model backbone}—and uncover two key insights:
(1) \textit{Latent-space fusion}, which integrates modality and structural cues at the feature level, yields more consistent and substantial gains than prompt-level fusion; and 
(2) \textit{VLM-based models} achieve the highest overall performance, underscoring their strong potential as backbones for multimodal graph learning tasks.

\item \textbf{Open-source benchmark library.} 
We release the GraphVLM benchmark, including datasets and codes, to facilitate future research in multimodal graph learning.
%We will release the GraphVLM benchmark, including datasets and codes, to foster future research and reproducibility in multimodal graph learning.
\end{itemize}

%\begin{itemize}[noitemsep,leftmargin=*,label=\ding{70}]
%\item \textbf{Comprehensive Benchmark}. GraphVLM provides a unified experimental framework for a fair comparison of state-of-the-art graph learning methods, including GNN-based, LLM-based, and VLM-based approaches, in six popular multimodal graph datasets. Notably, our benchmark highlights the most promising solution and offers guidance for future directions in multimodal graphs.
%\item \textbf{Multi-Dimensional Analysis}. We systematically evaluate existing graph methods from multiple perspectives, including domain-specific and structure-aware multimodal alignment, the integration of visual attributes via data augmentation, and task-specific supervised fine-tuning with or without structural knowledge. \textit{\textbf{Our Key Findings:}} \textit{i)} Without VLM enhancement, GraphLLMs outperform both GNN-based and VLM-based methods in MMGL tasks. \textit{ii)} VLM-as-Predictor approaches deliver the most significant performance gains, demonstrating their strong potential as the backbone model for MMGL. \textit{iii)} The sparse graph structures and imbalanced multimodal quality heavily influence the structure information effectiveness in MMGL.
%\item \textbf{Open-sourced benchmark library}. We have made our benchmark library publicly available on Github, aiming to facilitate future research endeavors. We have also outlined potential future directions based on our benchmark findings to inspire further investigations.
%\end{itemize}

% \section{Preliminary and Background}
\section{Formulations and Background}
\label{sec:preliminary}
In this section, we formally define multimodal graphs and then discuss three paradigms for multimodal graph learning.

\subsection{Multimodal Graph Definitions}
% (plain word description about MMGs: attribute level, each node with multimodal information)
We focus on text and image modalities, which are the most widely available and standardized sources of information for graph-based tasks. Formally, a multimodal graph (MMG) is defined as $\mathcal{G} = (\mathcal{V}, \mathcal{A}, \mathcal{T}, \mathcal{I}, \mathcal{Y})$, 
where $\mathcal{V}$ is the set of nodes, $\mathcal{A}$ the adjacency matrix, and $\mathcal{T}$ and $\mathcal{I}$ denote the textual and visual attributes of nodes, respectively. 
$\mathcal{Y}$ represents the label space for node-level prediction. 
In this work, we focus on the node classification task: given a node $v_i \in \mathcal{V}$ with a missing label $\mathcal{Y}_{v_i}$, the objective is to predict it based on both structural and multimodal information.

\subsection{Multimodal Graph Learning Methods}
\label{sec:graphmethods}
%Existing graph learning solutions can be broadly categorized into two main classes, GNN-based and LLM-based, and we explore VLMs as a third potential approach for multimodal graph learning (MMGL).

\vspace{3pt}
\noindent\textbf{GNN-based Methods.} 
Graph Neural Networks (GNNs) ~\cite{GNN, hamilton2017graphSAGE, mmgcn, tao2020mgat} are widely used in graph learning tasks due to their ability to aggregate information from neighboring nodes to learn expressive node embeddings. The update rule for the embedding  $\mathbf{h}_{v_i}^{(l)}$ of the node ${v_i}$ at GNN layer $l$ can be formulated as:
\begin{equation}
\mathbf{h}_{v_i}^{(l+1)} = \psi\left(\phi\left({\mathbf{h}_{{v_i}'}^{(l)} : {v_i}' \in \mathcal{N}({v_i})}\right), \mathbf{h}_{v_i}^{(l)}\right),
\end{equation}
where $\mathcal{N}(v_i)$ denotes the set of neighboring nodes of $v_i$, and $\phi(\cdot)$ and $\psi(\cdot)$ are aggregation and update functions, respectively. By stacking $L$ GNN layers, the final node embeddings can be utilized for downstream tasks. 

%\noindent\textbf{\textit{{\ding{228}} Challenge 1: Aligning Modalities with Structural Data.}}
%When GNN-based methods are applied to MMGL, they face the major challenge of modality alignment. Specifically, conventional GNN-based methods (e.g., GCN\cite{kipf2016gcn}, GAT\cite{velivckovic2017gat}, GraphSAGE\cite{hamilton2017graphSAGE}) primarily focus on aggregating structural information from graph topologies, often lacking the capability to effectively integrate and align heterogeneous modalities such as visual and textual data. An intuitive approach to address this limitation is to use modality-aligned pre-trained encoders (e.g., CLIP\cite{radford2021learning}, ImageBind\cite{girdhar2023imagebind}) to generate aligned visual and textual features as the input for GNN-based modeling. However, the question of how to effectively fuse both multimodal features and structural information remains an open research problem.

\vspace{3pt}
\noindent\textbf{LLM-based Methods (GraphLLMs).}
LLMs handle graph-based tasks by leveraging their language understanding to reason over structured data. GraphLLMs~\cite{ he2024unigraph, chen2024llaga, zhang2024graphtranslator} typically adopt a projection-based approach, mapping node embeddings into the LLM token space for unified processing. This process can be represented as:
\begin{equation}
\mathbf{z}_{v_i} = \text{Projector}(\mathbf{h}_{v_i}) \in \mathbb{R}^{d_\mathrm{LLM}},
\end{equation}where $\mathbf{h}_{v_i}$ is the GNN-generated node embedding, and $\mathbf{z}_{v_i}$ is the embedding projected into the LLM token space.
Alternatively, some GraphLLMs \cite{tang2024graphgpt, ye2023language} adopt instruction-based prompting methods that convert graph data into text sequences, enabling LLMs to infer graph-related insights through instruction-based learning. 
%Typically, instruction tuning is also applied for specific downstream tasks, enhancing performance and task alignment.

%(e.g., GraphGPT\cite{tang2024graphgpt}, UniGraph\cite{he2024unigraph})
%(e.g., LLaGA\cite{chen2024llaga}, InstructGLM\cite{ye2023language})
%\noindent\textbf{\textit{{\ding{228}} Challenge 2: Integrating Visual Modality into GraphLLMs.}} 
%A key challenge in applying GraphLLMs to MMGL is their inherent limitation in handling multimodal inputs, particularly visual information. Most existing GraphLLMs are designed for TAGs and rely solely on textual modalities, leading to sub-optimal performance when dealing with MMGs. The absence of image information restricts the model’s ability to capture visual-semantic correlations, resulting in incomplete feature representations and reduced effectiveness in downstream tasks. Nevertheless, it is unclear how to effectively enable the seamless integration of visual information into unimodal GraphLLMs.

\vspace{3pt}
\noindent\textbf{Vision Language Models (VLMs).}
Recent advances in VLMs enable a unified understanding of visual and textual modalities, providing new opportunities for multimodal graph learning. 
By leveraging their intrinsic cross-modal alignment capability, VLMs can jointly encode image--text pairs into a shared representation space. 
Formally, a VLM predicts a response $\mathcal{R}$ given an instruction prompt $\mathcal{P}$ and a multimodal input $\mathcal{M}$:
\begin{equation}
\mathcal{R} = \mathrm{VLM}(\mathcal{P}, \mathcal{M}; \theta),
\end{equation}
where $\theta$ denotes the model parameters. 
This unified formulation allows VLMs to integrate visual, textual, and structural information within graph-based reasoning frameworks.

%\noindent\textbf{\textit{{\ding{228}} Challenge 3: Enabling Structure Awareness in MMLMs.}} 
%While VLMs excel at aligning multimodal information, a critical challenge lies in awareness of the structural properties of graph data. With only instruction and modality inputs, VLMs often struggle to maintain structural awareness when processing graph data. A simple and lightweight approach to applying VLMs to MMGL tasks involves updating the instruction prompt with graph structure information $\mathcal{I}^G$. However, the efficiency of this strategy remains uncertain, and the effectiveness of VLMs as predictors for MMGL tasks is still an open question.

\section{Benchmark Design}
\label{sec:eval_frame}

In this section, we present the experimental design for the three categories of existing methods. Figure \ref{fig:graphVLM} illustrates the overall benchmark design together with a timeline of representative graph learning research.

\subsection{VLM-as-Encoder}
\label{sec:enhancer}

As introduced in Sec.~\ref{sec:graphmethods}, GNN-based methods take node embeddings as input and leverage graph structure to make predictions. 
In our benchmark, we explore how VLMs can serve as feature encoders for MMGL, examining whether stronger modality alignment and structure awareness at the embedding level can improve downstream performance. 

\noindent\textbf{Feature Encoders.}
To analyze the effects of multimodal alignment and structure-aware encoding, we employ pre-trained vision-language models (PVLMs) as node encoders and examine three representative variants:

\noindent{\textit{\ding{182} Pre-trained PVLM.}}
We first adopt the original PVLM without any task-specific tuning, using CLIP~\cite{clip} as a representative model and concatenating its text and image embeddings. 
To analyze the impact of modality configuration, we also compare text- and vision-only embeddings for the off-the-shelf multimodal encoder.

\noindent{\textit{\ding{183} Fine-tuned PVLM (PVLM-F).} 
To enable better modality alignment on graph data, we fine-tune the PVLM on each specific MMG dataset. 
Following the CLIP training paradigm, we employ a contrastive learning objective that maximizes the cosine similarity between visual and textual embeddings, enabling improved cross-modal consistency.

\noindent{\textit{\ding{184} Structure-aware PVLM (PVLM-F-S).} 
To further incorporate structural relationships among nodes, we extend the fine-tuned PVLM into a structure-aware variant. 
The model is jointly optimized within a GNN framework, where for each anchor node $v_i$, we randomly sample $m$ nodes from its 1-hop neighborhood to capture contextual dependencies. 
A structure-aware contrastive loss is applied to align multimodal features with the underlying graph topology:
\begin{equation}
    \mathcal{L}_v = -\log 
    \frac{\exp \left( \text{sim}(\mathcal{E}_{TI}^{v_i}, \mathcal{E}_{TI}^{v_j}) / \tau \right)}
    {\sum\limits_{v_k \in \mathcal{B}} 
    \exp \left( \text{sim}(\mathcal{E}_{TI}^{v_i}, \mathcal{E}_{TI}^{v_k}) / \tau \right)},
\end{equation}
where $\mathcal{E}_{TI}^{v_i}$ denotes the concatenated text–image embedding of the central node, $\mathcal{E}_{TI}^{v_j}$ is the embedding of its neighbors, $\tau$ is the temperature parameter, and $\mathcal{B}$ is the training batch.

\vspace{0.6mm}
\noindent\textbf{GNN Models.}
We experiment with several widely adopted GNN architectures, including GCN~\cite{kipf2016gcn}, GraphSAGE~\cite{hamilton2017graphSAGE}, MMGCN~\cite{mmgcn}, and MGAT~\cite{tao2020mgat}.  
We also include UniGraph2~\cite{uniGraph2}, a multimodal graph foundation model that captures both modality-specific and structural features. 
For completeness, a simple MLP~\cite{rosenblatt1958perceptron} is used as a baseline without structural integration. 
Training details and model settings are provided in Appendix~\ref{appendix:implementation}. 

\subsection{VLM-as-Aligner}
To explore how GraphLLMs can be adapted for MMGL, we design intuitive extensions that leverage VLMs as modality aligners. Specifically, we consider two complementary strategies, \textit{latent-space fusion} and \textit{prompt-level alignment}, which correspond to integrating multimodal signals at different stages of the model.

\noindent\textbf{{\ding{182} Latent-Space Aligner.} }
At the feature level, we enhance GraphLLMs by integrating multimodal embeddings in place of the original unimodal node representations, while preserving the original model architecture and training pipeline. This design allows the model to jointly reason over textual and visual features within the latent space, achieving multimodal understanding without altering the prompt structure.

\noindent\textbf{{\ding{183} Prompt-Level Aligner.} }
At the prompt level, we follow the paradigm of converting visual information into textual form to align modalities, as demonstrated in recent VLM works~\cite{minigpt4, guo2023imagestextualpromptszeroshot}. 
With this extension, we augment each node attribute with its corresponding visual description, enabling multimodal reasoning in a natural language format. We further consider a structure-aware variant that incorporates visual information from neighboring node attributes to enhance context modeling.

$\bullet$ \textit{Visual-augmented Node Prompt.}
For each node, the VLM converts its visual attribute $I$ into a generated textual description $\mathcal{T}^I$. 
Given GraphLLM’s limited context length, we summarize the original textual content $\mathcal{T}$ and the generated description $\mathcal{T}^I$ into a concise summary $\mathcal{T}^S$, whose embedding $\mathcal{E}_{\mathcal{T}^S}$ serves as the feature input. 
The process is formulated as:
\begin{equation}
\label{equation:generate}
\mathcal{T}^I = \mathrm{VLM}(\mathcal{P}_{\text{Generation}}, \mathcal{I}; \theta)
\end{equation}
\begin{equation}
\label{equation:summary}
\mathcal{T}^S = \mathrm{VLM}(\mathcal{P}_{\text{Summary}}, (\mathcal{T} + \mathcal{T}^I); \theta)
\end{equation}
where $\mathcal{P}_{\text{Generation}}$ and $\mathcal{P}_{\text{Summary}}$ denote task-specific instruction prompts, and $\theta$ denotes VLM parameters.

$\bullet$ \textit{Structure-aware Augmentation.}
To further integrate graph topology, we augment the visual information of both the anchor node and its neighbors. 
This yields a structure-aware multimodal summary $\mathcal{T}^{SS}$, whose embedding $\mathcal{E}_{\mathcal{T}^{SS}}$ serves as the feature input:
\begin{equation}
\label{equation:summary_neighbor}
\mathcal{T}^{SS} = \mathrm{VLM} (\mathcal{P}_{\text{Summary}}, (\mathcal{T} + \mathcal{T}^I + \sum_{v_j \in \mathcal{N}(v_i)} \mathcal{T}^I_{v_j}); \theta)
\end{equation}
where $\mathcal{N}(v_i)$ is the neighbor set of node $v_i$, and $\mathcal{T}^I_{v_j}$ is the image-derived description of node $v_j$. 
All prompts and implementation details are provided in Appendix~\ref{appendix:prompt}.

\vspace{0.6mm}
\noindent\textbf{GraphLLM Models.} 
We involve several representative GraphLLMs considering distinct architecture paradigms, including GraphPrompter~\cite{liu2024graphPrompter}, LLaGA~\cite{chen2024llaga}, GraphGPT~\cite{tang2024graphgpt}, GraphTranslator~\cite{zhang2024graphtranslator}, and state-of-the-art MLaGA~\cite{mlaga2025}. CLIP~\cite{clip} is employed as the latent-space aligner to provide multimodal embeddings, while Qwen-VL~\cite{bai2023qwenvlversatilevisionlanguagemodel} serves as the prompt-level VLM aligner for prompt augmentation.

\subsection{VLM-as-Predictor}
VLMs possess inherently strong multimodal alignment and reasoning capabilities, offering new opportunities to improve performance in MMGL. To further adapt them to graph-structured data, we perform supervised fine-tuning (SFT) under both structure-aware and non-structure-aware settings. In the structure-aware setting, we enhance the model’s ability to reason over both unimodal and multimodal neighbor node attributes through two complementary strategies:

\noindent\textbf{{\ding{182} Explicit Prompt-Level Fusion.}} 
We construct instruction prompts that explicitly incorporate the attributes of each anchor node along with its neighboring nodes. 
For structural information, we select the top-$k$ ($k$=3) most similar neighbors (by cosine similarity) from the anchor node’s $h$-hop ($h$=1) neighborhood and append their textual, visual, or multimodal attributes to the prompt. 
This design enables the model to directly contextualize the anchor node within its local structure during reasoning.

\noindent\textbf{{\ding{183} Implicit Latent-Space Fusion.}}  
To implicitly encode structural information, we aggregate the representations of neighbor nodes into unified \textit{structure-aware tokens} that are then injected into the model’s latent space.

$\bullet$ \textit{Visual Modality}. Following common visual feature aggregation practices~\cite{li2023blip2bootstrappinglanguageimagepretraining, kim2021viltvisionandlanguagetransformerconvolution, li2023llavamedtraininglargelanguageandvision}, we extract patch-level embeddings of neighbor images using the vision encoder and perform average pooling to obtain a compact latent representation summarizing the visual neighborhood context.  

$\bullet$ \textit{Textual Modality.} Inspired by the recent GraphLLM framework design~\cite{chen2024llaga, tang2024graphgpt}, we represent each node with a single semantic token derived from its textual attribute. Specifically, we average the final-layer token embeddings of the text encoder to obtain the node-level representation. All neighbor text tokens are then concatenated to form a unified textual neighborhood representation.  

These aggregated multimodal representations provide a fine-grained structural signal, enabling the VLM to reason jointly over multimodal node representations and their underlying structural dependencies.
Detailed implementation settings and prompt templates are provided in Appendices~\ref{appendix:implementation} and~\ref{appendix:prompt}.

\vspace{0.6mm}
\noindent \textbf{VLM Models.} 
For VLM-as-predictor approaches in MMGL tasks, we include three representative VLMs: LLaVA-1.5-7B~\cite{liu2023llava}, Qwen-VL-7B~\cite{bai2023qwenvlversatilevisionlanguagemodel}, and Qwen2.5-VL-7B~\cite{bai2025qwen25vltechnicalreport}. We strictly follow their official guidelines for LoRA fine-tuning.

\section{Experiments}
\label{sec:exp_analysis}
In this section, we conduct extensive experiments to better understand the role of VLMs in MMGL by addressing the following research questions (\textbf{RQs}):
\textbf{\textit{RQ1:}} How effective is VLM-as-Encoder in enhancing traditional GNN-based methods with multimodal and structural cues? 
\textbf{\textit{RQ2:}} How does VLM-as-Aligner facilitate multimodal reasoning in GraphLLMs through prompt- and latent-level alignment? 
\textbf{\textit{RQ3:}} Can VLMs fine-tuned on graph-structured data serve as standalone predictors for multimodal graph learning tasks? 
\textbf{\textit{RQ4:}} Which functional role of VLMs has the greatest potential to improve performance in MMGL?

%\noindent\textbf{\textit{{\ding{228}} RQ1:}} 

\subsection{Datasets}
This benchmark covers six different datasets from the Amazon co-purchase networks~\cite{jure2014snap, zhu2024multimodalgraphbenchmark} and the social network Reddit platform~\cite{hamilton2017graphSAGE, yan2024graph}. They vary in scale and density, with a wide range of the number of nodes and different average degrees. Each node represents a product or a post and is associated with multimodal information: textual description and visual image. Edges between nodes indicate the relationships of co-purchased or co-commented, reflecting product or user interactions. The labels for these nodes are categorical classes, representing different product categories or post types. We adopt a split setting of 60\% for training, 20\% for validation, and 20\% for testing. More detailed dataset statistics are shown in the Appendix~\ref{appendix:dataset}.

%~\ref{table:dataset}.
% More details on the implementations can be found in Appendix~\ref{appendix:implementation}

%\cite{yan2024graph}.

%%%%%%%%%%%%%%%%%%%%%%%
%%%%%%% Enhancer
%%%%%%%%%%%%%%%%%%%%%%%

\subsection{Impact of VLM-as-Encoder (RQ1)}
We conduct experiments using both unimodal and multimodal inputs across various GNN-based models and evaluate different encoders with multimodal inputs.

\begin{table*}[t]
\centering
\scriptsize

\resizebox{\textwidth}{!}{
\begin{tabular}{@{}l|l|cccccc@{}}
\toprule
\midrule
\textbf{Model} & \textbf{Encoder} & \textbf{Movies} & \textbf{Toys} & \textbf{Grocery} & \textbf{Arts} & \textbf{CDs} & \textbf{Reddit} \\ 
\midrule
\midrule

\multirow{3}{*}{MLP} 
& Imagebind & 40.14 {\tiny$\pm$0.14} & 52.30 {\tiny$\pm$0.45} &  65.17 {\tiny$\pm$0.17} & 74.79 {\tiny$\pm$0.37} & 30.76 {\tiny$\pm$0.90} &  42.62 {\tiny$\pm$0.27} \\
& CLIP &  45.12 {\tiny$\pm$0.12}   & \textbf{74.59} {\tiny$\pm$0.09} & \textbf{84.67} {\tiny$\pm$0.04} & 84.28 {\tiny$\pm$0.17} & \textbf{52.22} {\tiny$\pm$0.05} & 76.70 {\tiny$\pm$0.25} \\
& CLIP-F & \textbf{47.46} {\tiny$\pm$0.08} & 73.22 {\tiny$\pm$0.09} & 82.79 {\tiny$\pm$0.05} & 82.66 {\tiny$\pm$0.07} & 48.74 {\tiny$\pm$0.07} & \textbf{79.77} {\tiny$\pm$0.10} \\
& CLIP-F-S & 39.73 {\tiny$\pm$0.12} & 59.88 {\tiny$\pm$0.25} & 76.94 {\tiny$\pm$0.03} & \textbf{84.89} {\tiny$\pm$0.06} & 49.89 {\tiny$\pm$0.07} & 79.71 {\tiny$\pm$0.18} \\
\midrule

\multirow{3}{*}{GCN} 
& Imagebind & 45.64 {\tiny$\pm$0.56} & 71.02 {\tiny$\pm$0.13}  & 79.01 {\tiny$\pm$0.07} & 75.01 {\tiny$\pm$0.16} & 47.42 {\tiny$\pm$0.43} & \textbf{70.89} {\tiny$\pm$0.20}\\
& CLIP & 46.97 {\tiny$\pm$0.24} & \textbf{74.36} {\tiny$\pm$0.07} & 79.59 {\tiny$\pm$0.29} & \textbf{76.76} {\tiny$\pm$0.21} & \textbf{52.68} {\tiny$\pm$0.03} & 66.06 {\tiny$\pm$0.11} \\
& CLIP-F & \cellcolor{myblue} \textbf{47.88} {\tiny$\pm$0.24} & 74.29 {\tiny$\pm$0.07} & \textbf{79.67} {\tiny$\pm$0.03} & 76.52 {\tiny$\pm$0.09} & 51.47 {\tiny$\pm$0.21} & 67.67 {\tiny$\pm$0.13} \\
& CLIP-F-S & 46.72 {\tiny$\pm$0.24}  & 71.41 {\tiny$\pm$0.14}  & 79.19 {\tiny$\pm$0.13} & 76.75 {\tiny$\pm$0.10} & 52.58 {\tiny$\pm$0.09} & 67.77 {\tiny$\pm$0.18} \\
\midrule

\multirow{3}{*}{GraphSAGE} 
& Imagebind & 38.82 {\tiny$\pm$0.48} & 62.08 {\tiny$\pm$0.99} & 78.66 {\tiny$\pm$0.74} & 78.92 {\tiny$\pm$0.84}  & 27.30 {\tiny$\pm$3.98} & 62.66{\tiny$\pm$0.91} \\
& CLIP &  44.08 {\tiny$\pm$0.37}   & \cellcolor{myblue} \textbf{77.77} {\tiny$\pm$0.70} & \cellcolor{myblue} \textbf{86.05} {\tiny$\pm$0.24} & 85.35 {\tiny$\pm$0.34} & \textbf{54.75} {\tiny$\pm$0.08} & 76.48 {\tiny$\pm$0.30} \\
& CLIP-F & \textbf{46.08} {\tiny$\pm$0.38} & 76.75 {\tiny$\pm$0.70} & 85.60 {\tiny$\pm$0.08} & 84.85 {\tiny$\pm$0.13} & 53.03 {\tiny$\pm$0.06} & \textbf{80.03} {\tiny$\pm$0.02} \\
& CLIP-F-S & 44.92 {\tiny$\pm$0.38} & 73.44 {\tiny$\pm$0.23} & 83.84 {\tiny$\pm$0.08} & \textbf{86.97} {\tiny$\pm$0.03} & 54.11 {\tiny$\pm$0.36}  & 79.94 {\tiny$\pm$0.24} \\
\midrule

\multirow{3}{*}{MMGCN} 
& Imagebind & 44.99 {\tiny$\pm$0.87} & 74.06 {\tiny$\pm$0.18}  & 81.59 {\tiny$\pm$0.07} & 86.30 {\tiny$\pm$0.29} & 47.21 {\tiny$\pm$0.20} & 71.06 {\tiny$\pm$0.24} \\
& CLIP & \textbf{45.90} {\tiny$\pm$0.71} & \textbf{75.36} {\tiny$\pm$0.43} & \textbf{84.63} {\tiny$\pm$0.38} & \cellcolor{myblue} \textbf{88.92} {\tiny$\pm$0.33} & 51.33 {\tiny$\pm$0.61} & 80.99 {\tiny$\pm$0.24} \\
& CLIP-F & 45.42 {\tiny$\pm$1.53} & 75.22 {\tiny$\pm$0.37} & 83.61 {\tiny$\pm$0.61} & 88.22 {\tiny$\pm$0.86}  & 50.61 {\tiny$\pm$0.73} & 81.71 {\tiny$\pm$0.22} \\
& CLIP-F-S & 44.87 {\tiny$\pm$1.66} & 72.74 {\tiny$\pm$0.26} & 81.99 {\tiny$\pm$0.41} & 87.78 {\tiny$\pm$0.28} & \textbf{52.18} {\tiny$\pm$0.40} & \textbf{81.71} {\tiny$\pm$0.31} \\
\midrule

\multirow{3}{*}{MGAT} 
& Imagebind & 45.51 {\tiny$\pm$0.33}  & 72.32 {\tiny$\pm$1.75} & 83.94 {\tiny$\pm$0.36} & 87.82 {\tiny$\pm$0.35} & 50.66 {\tiny$\pm$0.48} & 74.36 {\tiny$\pm$0.83} \\
& CLIP & \textbf{46.33} {\tiny$\pm$0.90} & 74.59 {\tiny$\pm$0.70}  & 82.98 {\tiny$\pm$1.20}  & 88.35 {\tiny$\pm$0.23} & 53.48 {\tiny$\pm$0.49}  & 78.87 {\tiny$\pm$1.32}  \\
& CLIP-F & 46.10 {\tiny$\pm$0.54} &	\textbf{75.52} {\tiny$\pm$0.81}  & \textbf{85.29} {\tiny$\pm$0.06} & \textbf{88.75} {\tiny$\pm$0.43}  & 54.29 {\tiny$\pm$0.62} & 81.33 {\tiny$\pm$0.36} \\
& CLIP-F-S & 44.74 {\tiny$\pm$0.97}  & 73.04 {\tiny$\pm$1.24}  & 82.98 {\tiny$\pm$1.20}  & 88.48 {\tiny$\pm$0.25} & \cellcolor{myblue} \textbf{55.70} {\tiny$\pm$0.49}  & \cellcolor{myblue} \textbf{81.74} {\tiny$\pm$0.34} \\
\midrule

\multirow{1}{*}{UniGraph2}       
& CLIP + GNN & 45.91 & 72.70 & 80.14 & 78.81 & 52.13 & 70.64 \\
\midrule
\bottomrule

\end{tabular}
}
\caption{Node classification accuracy (\%) comparison under different enhanced encoder settings. 
\textbf{Bold} values denote the highest performance within the same model and dataset setup. \colorbox{myblue}{Blue} cells indicate the highest result within each dataset.}
\label{table:encoder}
\end{table*}

\noindent\textbf{\textit{\ding{70} Finding 1:} Multimodal inputs consistently boost GNN-based model performance over unimodal inputs.}

As shown in Figure~\ref{fig:modality}, introducing multimodal inputs (\textit{text + image}) consistently enhances the performance of all GNN baselines compared with unimodal settings. Across models, image-only inputs underperform text-only ones, yet their fusion through embedding integration yields substantial gains. In particular, for the MLP model, multimodal input improves accuracy by \textbf{\textit{+5.18\%}} over text-only and \textbf{\textit{+13.61\%}} over image-only, highlighting the importance of multimodal representations in capturing complementary information.

\noindent\textbf{\textit{\ding{70} Finding 2:} Existing modality–structure alignment strategies remain insufficient for enabling GNN-based models to fully exploit multimodal graph information.}

From Table~\ref{table:encoder}, we observe that the structure-aware encoder \textit{CLIP-F-S} improves node classification accuracy for MLP, GraphSAGE, MMGCN, and MGAT on the Arts and CDs datasets. However, the overall best results across datasets are achieved by GCN, GraphSAGE, and MMGCN with the \textit{CLIP} and \textit{CLIP-F} encoders—both lacking explicit structure awareness. This suggests that incorporating structural information into a single embedding is insufficient to jointly align multimodal content and graph structure. Interestingly, on the Reddit dataset, where visual quality is high but textual content is sparse, all modality-enhanced encoders yield clear performance gains, suggesting that vision information is particularly beneficial when text is limited. These observations indicate that simply introducing structure-aware fusion is not sufficient. Future progress in MMGL requires more robust alignment mechanisms that jointly optimize feature fusion and graph-structural coherence to fully leverage multimodal signals. 

In conclusion, PVLMs yield notable performance gains in multimodal settings compared to unimodal ones, consistent with prior benchmarks~\cite{zhu2024multimodalgraphbenchmark, yan2024graph}. 
However, such improvements mainly capture shallow correlations between semantic and visual cues rather than structured dependencies critical for graph-based learning. 
When fine-tuning is applied, either with or without structure awareness, the performance of GNN-based methods improves, validating our motivation to further explore the role of VLMs as encoders for MMGL.

%From Table~\ref{table:encoder_comparision}, we can observe that the structure-aware encoder \textit{CLIP-F-S} does not improve node classification accuracy for MLP, GCN, and GraphSAGE models on the Movies and Arts datasets. With the same combined text and image multimodal inputs, \textit{Enhancer-v1} achieves gains of 2.25\%, 0.91\%, and 2\% over \textit{Enhancer-v0} on the Movies dataset, while \textit{Enhancer-v2} shows improvements of 0.61\% and 1.62\% on the CDs dataset. However, these intuitive alignment methods, including standalone multimodal training with the SOTA encoder CLIP and contrastive learning with structural knowledge, fail to consistently deliver stable performance gains across diverse datasets. This highlights the need for advanced alignment techniques that can more effectively integrate multiple node attribute modalities with graph structural information.

\begin{figure}[t]
    \centering
    \includegraphics[width=0.48\textwidth]{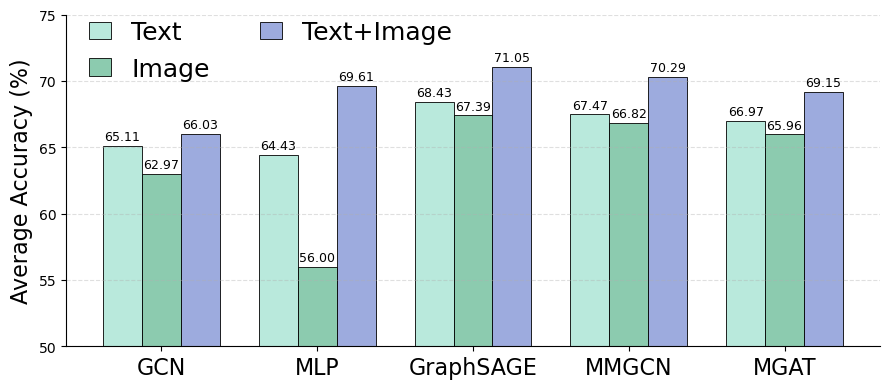}
    \caption{
        Average accuracy (\%) across six datasets under different modalities for GNN-based methods, using CLIP as the encoder.
    }
    \label{fig:modality}
    \vspace{-2mm}
\end{figure}

%%%%%%%%%%%%%%%%%%%%%%%
%%%%%%% Augmenter
%%%%%%%%%%%%%%%%%%%%%%%

\subsection{Adaptation of VLM-as-Aligner (RQ2)}
We compare the performance of original text-only inputs with multimodal extensions where VLMs serve as modality aligners for GraphLLMs. Specifically, we explore two alignment strategies: (1) \textit{prompt-level} augments the textual prompt with image-derived descriptions, and (2) \textit{latent-space}, which concatenates image and text embeddings as multimodal feature inputs for GraphLLMs. These strategies aim to assess whether integrating modality information at the prompt or embedding level benefits multimodal graph learning. Results are summarized in Table~\ref{table:aligner}, showing the performance with different alignment paradigms under various model architectures.

%%%%%%%%%%%%%%%%%%%%%%%
%%%%%%% Augmenter Table
%%%%%%%%%%%%%%%%%%%%%%%

\begin{table}[htbp]
    \centering
    \resizebox{0.5\textwidth}{!}{
    \begin{tabular}{l@{\hspace{2mm}}|l|cccccc}
    \toprule
    \midrule
    \textbf{Model} & \textbf{Input} & \textbf{Movies} & \textbf{Toys} & \textbf{Grocery} & \textbf{Arts} & \textbf{CDs} \\ 
    \midrule
    \midrule
    
    \multirow{4}{*}{GraphPrompter}       
    &  \textsymb & \textbf{46.36} & 74.08 & 85.00 & 84.41 & 46.27  \\
    & \textsymb ~\promptfuse ~~\imagesymb  & 45.55 & \textbf{76.56} & \textbf{85.47} & 64.87 & 52.06  \\
    & \textsymb ~\promptfuseS ~\imagesymb & 46.12 & 75.07 & 80.57 & 56.61 & \textbf{52.58} \\
    & \textsymb ~\latentfuse ~~\imagesymb  & 45.88 & 70.96 & 84.91 & \textbf{85.85} & 45.73 \\

    \midrule
    
    \multirow{4}{*}{LLaGA}       
    & \textsymb  & 49.57 & 77.51 &  86.25 & \textbf{89.32} & 53.12 \\
    & \textsymb ~\promptfuse ~~\imagesymb & 48.73 & 74.97 & 86.16 & 76.34 & 54.45  \\
    & \textsymb ~\promptfuseS ~\imagesymb & 45.79 & 77.27 & 85.36 & 70.85 & 55.42 \\
    & \textsymb ~\latentfuse ~~\imagesymb  & \cellcolor{myblue}  \textbf{50.61} & \textbf{79.34} & \cellcolor{myblue} \textbf{86.83} & 88.83 & \cellcolor{myblue}  \textbf{56.29} \\

    \midrule
    
    \multirow{2}{*}{GraphGPT}       
    & \textsymb  & \textbf{10.19} & \textbf{37.38} & 59.76 & 57.35 & 23.86 \\
    & \textsymb ~\promptfuse ~~\imagesymb & 8.82 & 36.10 & \textbf{59.87} & \textbf{58.73} & \textbf{29.57}  \\
    %& \textsymb ~\latentfuse ~~\imagesymb  & 8.82 &  &  & 58.73 & 29.57 \\

    \midrule
    
    \multirow{2}{*}{GraphTranslator}       
    & \textsymb  & \textbf{7.11} & 11.96 & \textbf{16.28} & \textbf{33.00} & 9.35 \\
    & \textsymb ~\promptfuse ~~\imagesymb & 5.34 & \textbf{19.96} & 13.68 & 13.90 & \textbf{9.53} \\
    
    \midrule
    
    \multirow{1}{*}{MLaGA}       
    & \textsymb ~\latentfuse ~~\imagesymb & \textbf{49.78} & \cellcolor{myblue} \textbf{80.00} &  \textbf{85.78} & \cellcolor{myblue} \textbf{89.79} & \textbf{54.45} \\

    \midrule
    \bottomrule
    \end{tabular}
    }
    \caption{Node classification accuracy (\%) of GraphLLM-based methods under different input and alignment strategies. \textsymb~indicates the text modality, while~~\imagesymb~represents the vision modality. \promptfuse~indicates prompt-level alignment, \promptfuseS~represents structure-aware prompt-level alignment, and~\latentfuse~denotes latent-space alignment. \textbf{Bold} denotes the highest performance within the same model and dataset setup. \colorbox{myblue}{Blue} cells indicate the highest result within each dataset.} 
    \label{table:aligner}
    \vspace{-5mm}
\end{table}

%\begin{figure}[t]
%    \centering
%    \includegraphics[width=0.36\textwidth]{sec/figures/augmentor_performance.png}
%    \captionof{figure}{Different aligner augmentation methods comparison}
%    \label{fig:augmentor}
%\end{figure}

\noindent\textbf{\textit{\ding{70} Finding 3. 
%Incorporating modality aligners consistently enhances multimodal graph learning. 
VLMs are effective modality aligners for extending GraphLLMs to multimodal graph learning.}}

Incorporating modality aligners consistently improves multimodal graph learning performance of standard GraphLLM methods.
As shown in Table~\ref{table:aligner}, GraphPrompter, LLaGA, and GraphGPT achieve their best results on \textit{4}, \textit{4}, and \textit{3} out of \textit{5} datasets, respectively, when using multimodal inputs with either prompt-level or latent-space aligner. 
Moreover, the top performance of all datasets appears in multimodal baselines, highlighting that integrating visual and textual features through the modality aligner effectively enhances graph reasoning.
These results demonstrate that GraphLLMs can be readily extended to multimodal graph learning by leveraging VLMs as modality aligners that fuse heterogeneous signals into graph-structured representations.
%This demonstrates that GraphLLMs can leverage VLMs as \textit{modality aligners} to fuse heterogeneous information into graph-structured representations, thereby extending their capability to multimodal graph learning tasks.

\vspace{0.25mm}
\noindent\textbf{\textit{\ding{70} Finding 4. In the VLM-as-Aligner paradigm, Latent-space alignment enables more effective multimodal fusion than prompt-level alignment. }}

Across all settings, latent-space alignment consistently outperforms prompt-level alignment.
The highest accuracies in Movies (\textit{50.61\%}), Grocery (\textit{86.83\%}), and CDs (\textit{56.29\%}) are achieved by LLaGA with latent-space alignment, while Toys (\textit{80.00\%}) and Arts (\textit{89.79\%}) peak under MLaGA. This improvement stems from the fact that latent-space alignment allows multimodal embeddings to be further aligned through learnable projectors in LLaGA or MLaGA during training, enabling the model to better integrate structural cues and reduce the noise or redundancy introduced by direct text augmentation at the prompt level.

Regarding prompt-level alignment, whether structure-aware or not, no clear trend suggests a significant advantage.
This may stem from the information bottleneck caused by converting multimodal signals into a single textual channel: explicit transformation can cause visual feature loss or noise injection.
In contrast, directly utilizing raw image embeddings allows GraphLLMs to retain richer vision semantics, facilitating more effective learning during both projector training and instruction tuning.

%%%%%%%%%%%%%%%%%%%%%%%
%%%%%%% Predictor Table
%%%%%%%%%%%%%%%%%%%%%%%

\begin{table}[htbp] 
\centering

\resizebox{0.5\textwidth}{!}{

\begin{tabular}{@{}l|ll|cccccc@{}}
\toprule
\midrule
\textbf{VLM} & \textbf{SFT} & \textbf{Fusion}  & \textbf{Movies} & \textbf{Toys} & \textbf{Grocery} & \textbf{Arts} & \textbf{CDs} & \textbf{Reddit} \\ 
\midrule
\midrule

\multirow{3}{*}{LLaVA-1.5-7B} 

& \nomark &  &  13.19 & 47.52 & 57.88 & 72.96 & 41.30 & 34.82\\
%& No & Neighbor Text+Vision & \underline{13.19} & \underline{47.52} & \underline{57.88} & \underline{72.96} & \underline{41.30}  \\
&  \yesmark &  &  \textbf{45.58} &  74.32 &  85.89 &  \textbf{84.47} &  53.55 &  82.45\\
&  \yesmark &  \textsymb ~ \promptfuse & 44.65 &  \textbf{74.82} & \textbf{85.91} &  82.94 &  \textbf{53.73} &  \textbf{82.76} \\
%& Yes & Neighbor Image &  &  &  &  & \\
%& Yes & Neighbor Text+Vision &  &  &  &  & \\ 

\midrule

\multirow{8}{*}{Qwen2.5-VL-7B} 

& \nomark & & 14.96 & 58.27 & 68.79 & 69.34 & 34.94 & 34.33 \\
&  \yesmark & & \cellcolor{myblue} \textbf{54.21} & \cellcolor{myblue} \textbf{82.56} & 88.32 & 93.37 & 54.86 & 67.48 \\
&  \yesmark &  \textsymb ~\promptfuse & 53.10 & 80.38 & \cellcolor{myblue} \textbf{88.53} & 92.43	& 54.13  & 67.04 \\
&  \yesmark &  \imagesymb ~\promptfuse & 53.43 & 80.00 & 88.48 & \cellcolor{myblue} \textbf{93.62} & 55.19 & \textbf{70.01} \\
&  \yesmark &  \textsymb~\imagesymb ~\promptfuse & 53.52 & 80.77 &  88.36 & 93.56 & \textbf{55.34} & 69.13 \\

\midrule

\multirow{8}{*}{Qwen-VL-7B} 

& \nomark &  & 12.56 & 48.90 & 43.73 & 67.69 & 37.51 & 47.52 \\
%& \nomark & \textsymb~\imagesymb & \promptfuse &  29.18 &  57.04 &  66.21 &  68.66  &  41.64 &  56.05 \\
&  \yesmark &  & \textbf{54.18} &  80.38  &  88.33  &  92.43 &  58.99  &  85.76 \\
&  \yesmark &  \textsymb ~\promptfuse & 51.66  &  80.96 &  \underline{88.34} &  92.04  &  \underline{58.81}  &  \underline{85.96} \\
&  \yesmark &  \imagesymb ~\promptfuse & 46.27  &  80.12  &  87.78 &  91.13 &  57.45  &  84.97\\
&  \yesmark &  \textsymb~\imagesymb ~\promptfuse & 45.88  &  79.80  &  87.78 &  91.15  &  57.01  &  85.25 \\ 
&  \yesmark & \textsymb ~\latentfuse & \underline{53.10} & \textbf{\underline{81.06}} & 88.29 & \underline{92.32} & 58.66 & 85.79\\
&  \yesmark &  \imagesymb ~\latentfuse & \underline{51.81} & \underline{80.74} & \textbf{\underline{88.49}} & \underline{92.48} & \underline{58.65} & \underline{86.09}\\
&  \yesmark &  \textsymb~\imagesymb ~\latentfuse &  \underline{48.49} & \textbf{\underline{81.06}} & \underline{88.04} & \textbf{\underline{92.66}} & \cellcolor{myblue} \textbf{\underline{59.82}} & \cellcolor{myblue} \textbf{\underline{86.67}}\\ 

%% & v3 & Neighbor Text & 52.77 & 80.19 & 88.17 & 91.82 & \textbf{59.15} \\
%& Yes & Neighbor Text & 52.77 ({\scriptsize $\uparrow$40.21\%}) & 80.19 ({\scriptsize $\uparrow$31.29\%}) & 88.17 ({\scriptsize $\uparrow$44.44\%}) & 91.82 ({\scriptsize $\uparrow$24.13\%}) & \textbf{59.15 ({\scriptsize $\uparrow$21.64\%})} \\

%& Yes & Neighbor Image & 50.16 ({\scriptsize $\uparrow$37.60\%}) & 78.84 ({\scriptsize $\uparrow$29.94\%}) & 87.31 ({\scriptsize $\uparrow$43.58\%}) & 91.15 ({\scriptsize $\uparrow$23.46\%}) & 58.00 ({\scriptsize $\uparrow$20.49\%}) \\

%& Yes & Neighbor Text+Vision & 50.67 ({\scriptsize $\uparrow$38.11\%}) & 78.86 ({\scriptsize $\uparrow$29.96\%}) & 87.55 ({\scriptsize $\uparrow$43.82\%}) & 91.84 ({\scriptsize $\uparrow$24.15\%}) & 58.04 ({\scriptsize $\uparrow$20.53\%}) \\

\midrule
\bottomrule
\end{tabular}
}

\caption{Node classification accuracy (\%) comparison under different VLM models, structure data modality, and structural fusion settings. 
Models are evaluated with and without SFT.
\promptfuse~indicates prompt-level structural fusion, and~\latentfuse~denotes latent-space structural fusion. 
\textbf{Bold} values mark the best performance within each model setup, while 
\colorbox{myblue!50}{\textbf{Blue}} cells highlight the overall best results across all baseline settings for each dataset. \underline{Underline} indicates the higher performance of fusion approaches under the same neighbor modality for each dataset in Qwen-VL-7B.} 

\label{table:predictor} 

% \vspace{-3pt}
\end{table}

\begin{figure}[t]
    \centering
    \includegraphics[width=0.45\textwidth]{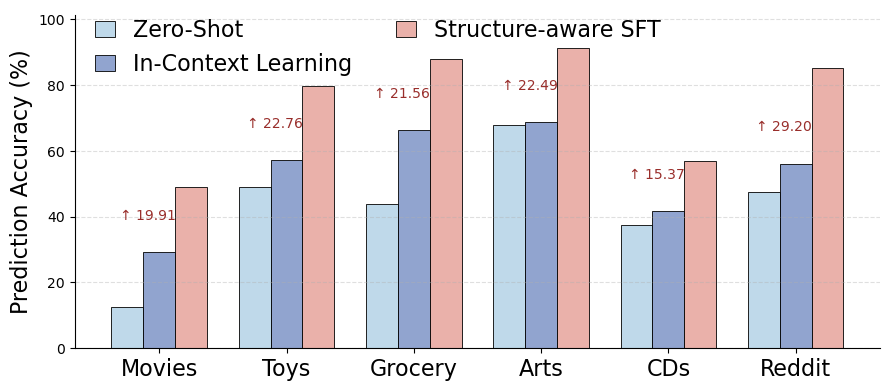}
    \caption{
        The impact of different VLM-based methods with multimodal structure information at the prompt level for node classification tasks, using Qwen-VL-7B as the backbone.
    }
    \label{fig:incontext-sft}
    \vspace{-1mm}
\end{figure}

%%%%%%%%%%%%%%%%%%%%%%%
%%%%%%% Predictor
%%%%%%%%%%%%%%%%%%%%%%%

\subsection{Effectiveness of VLM-as-Predictor (RQ3)}

Figure~\ref{fig:incontext-sft} and Table~\ref{table:predictor} present the results for zero-shot, in-context learning, and fine-tuned prediction methods. LLaVA-1.5-7B supports only a single image as input, so we conduct more structure-aware experiments using the Qwen-VL-7B model, and Qwen2.5-VL-7B is utilized as a stronger VLM baseline for reference. The following are two key findings:

\noindent\textbf{\textit{\ding{70} Finding 5:} Fine-tuning VLM could significantly improve its performance compared to in-context learning.} 

As shown in Figure~\ref{fig:incontext-sft}, supervised fine-tuning consistently outperforms both zero-shot and in-context learning (ICL) across all datasets, with the best performance achieved in every case under the SFT setting. Under zero-shot settings without structural information or fine-tuning, all VLMs achieve relatively low accuracy. However, fine-tuning, even without explicit structural inputs, yields dramatic gains (Table~\ref{table:predictor}). For instance, with Qwen-VL-7B on the \textit{Movies} dataset, accuracy rises from only 12.56\% to 54.18\%, achieving a remarkable \textbf{\textit{4.3$\times$}} improvement. LLaVA-1.5-7B demonstrates a similar trend with a \textbf{\textit{3.5$\times$}} gain. Specifically, Qwen-VL-7B achieves absolute improvements of \textbf{\textit{+21.48\%}} - \textbf{\textit{+44.61\%}} across all datasets.

We further compare \textit{structure-aware} ICL with \textit{structure-aware} SFT, where both incorporate multimodal neighbor information (text and image) as structural cues. While ICL provides a lightweight approach of introducing structure at the prompt level, SFT delivers much stronger adaptation, with additional accuracy gains of \textbf{\textit{+15.37\%}} - \textbf{\textit{+29.20\%}} across all datasets. This substantial gap indicates that fine-tuning enables VLMs to internalize structural dependencies more effectively than prompt-based context learning, thereby unleashing their full potential as general backbones for multimodal graph learning.

\noindent\textbf{\textit{\ding{70} Finding 6: In the VLM-as-Predictor paradigm, structure-aware fine-tuning and latent-space fusion deliver significant performance gains.}}

As shown in Table~\ref{table:predictor}, incorporating structural information during supervised fine-tuning consistently improves accuracy across multiple backbones, boosting performance on \textit{5 out of 6} datasets for Qwen-VL-7B and on \textit{4 out of 6} datasets for both LLaVA-1.5-7B and Qwen2.5-VL-7B. This confirms that integrating graph-aware cues during fine-tuning enables VLMs to more effectively leverage multimodal dependencies for node-level reasoning.

We also compare prompt-level and latent-space fusion under identical multimodal neighbor inputs based on the Qwen-VL-7B model. Latent-space fusion surpasses prompt-level fusion in \textbf{\textit{15 out of 18}} cases, yielding more stable convergence and higher accuracy by embedding structural cues directly within the model’s representation space. We attribute this to the fact that prompt-level fusion, which is performed through direct text concatenation, is constrained by limited context length and token interference, whereas latent-space fusion embeds structural information directly into the model’s internal representation space, allowing fine-tuning to more effectively capture structure cues.  

Overall, these results demonstrate that fine-tuned VLMs, when used as backbones, are highly effective for multimodal graph learning, and that latent-space fusion offers a practical and powerful way to incorporate structural information for further performance gains.

%%%%%%%%%%%%%%%%%%%%%%%
%%%%%%% Mixture-of-All
%%%%%%%%%%%%%%%%%%%%%%%

\subsection{Comparative Analysis of VLM Roles (RQ4)}
After evaluating each MMGL paradigm and the respective models, we combine all the paradigms to comprehensively assess and summarize the potential of VLMs for MMGL.

We compare the best performance of the original methods with VLM-enhanced models, highlighting the performance gains achieved through VLM integration. The results are illustrated in Figure~\ref{fig:bestperformance}. 
To evaluate the effectiveness of structure awareness across all approaches, we assess the extra performance gain provided by structure-aware methods in Figure~\ref{fig:structuregain}. This performance gain is calculated as the difference between the highest accuracy achieved by structure-aware methods and that achieved by structure-agnostic methods.
 
%%%%%%%%%%%%%%%%%%%%%%%
%%%%%%% Mix All Figures
%%%%%%%%%%%%%%%%%%%%%%%

\begin{figure}[t]
    \centering
    \includegraphics[width=0.95\linewidth]{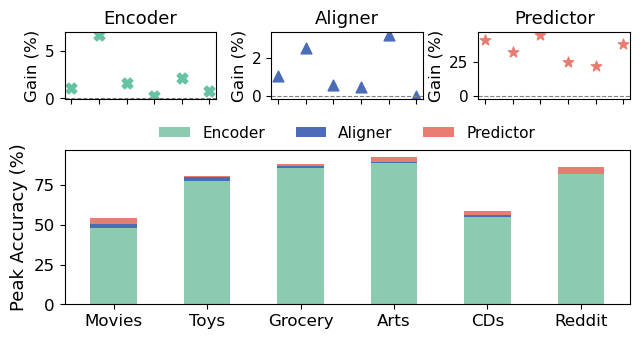}
    \caption{Performance comparison of three VLM roles.}
    \label{fig:bestperformance}
    \vspace{-1mm}
\end{figure}

\begin{figure}[t]
    \centering
    \includegraphics[width=0.44\textwidth]{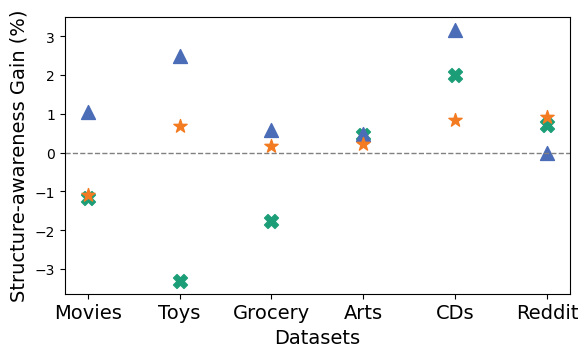}
    \caption{Extra performance gain of structure-awareness.}
    \label{fig:structuregain}
    \vspace{-2mm}
\end{figure}

\noindent\textbf{\textit{\ding{70} Finding 7: VLM-as-Predictor emerges as the most powerful paradigm for MMGL, while VLM-as-Encoder and VLM-as-Aligner generally provide meaningful gains.}}

As shown in Figure \ref{fig:bestperformance}, all three VLM roles bring benefits by introducing multimodal information for vanilla unimodal graph learning methods, demonstrating the overall improvement in MMGL. Among them, the \textit{VLM-as-Predictor} paradigm exhibits the most remarkable improvement: fine-tuning structure-aware VLMs transforms the lowest-performing zero-shot baseline into the top performer across all datasets, achieving the largest performance gap observed. This dramatic boost highlights that once structural cues are incorporated through supervised fine-tuning, VLMs can effectively internalize graph-aware multimodal reasoning. Overall, these results reveal that fine-tuned VLMs, when used directly as backbones, hold tremendous promise as a new foundation for multimodal graph learning.

\noindent\textbf{\textit{\ding{70} Finding 8: Structure cues benefit all three VLM roles, though their effect on VLM-as-Encoder remains unstable.}}

As shown in Figure~\ref{fig:structuregain}, incorporating structural information in the \textit{VLM-as-Predictor} and \textit{VLM-as-Aligner} paradigms yields consistent and reliable performance gains in almost all datasets.
In contrast, structure-aware fine-tuned \textit{Encoders} show limited stability, yielding improvements on only three datasets while degrading on others. 
This can be attributed to the fact that structure-aware predictors leverage fine-grained token sequences that can facilitate selective attention over neighbors and modalities in VLM backbones. Conversely, encoder-based approaches often rely on a single compressed embedding to integrate heterogeneous information, which introduces a representation bottleneck for complex neighborhood structures. % This can be attributed to low-level feature interference and the inherent difficulty of integrating heterogeneous modalities prior to high-level semantic abstraction.
Furthermore, results show that datasets with higher graph density, such as CDs whose average degree is as high as 47, benefit the most from structure awareness, whereas sparse graphs such as Toys and Grocery often show unstable or even negative gains.
These findings highlight the complexity of structure–modality interactions and suggest that future multimodal graph learning should build on strong VLM backbones, selectively injecting structural cues to enhance relational reasoning.

% Both graph structure density and modality quality significantly affect the effectiveness of structure awareness. Results reveal that for datasets with higher average degrees, particularly in the CDs dataset with a high average degree of 47, structure awareness consistently improves performance across all three approaches. In contrast, in sparse graphs such as Toys, structure awareness not only fails to provide benefits but may even degrade performance. Nevertheless, the modality aligner helps enrich node representations, thereby improving predictive performance. Additionally, the Movies dataset, with the lowest image quality (see Table~\ref{table:modality}), demonstrates how poor image features can undermine the structural effectiveness of all methods. These findings highlight the need to consider sparse graph structures and imbalanced multimodal quality when dealing with MMGL.

\section{Conclusion}
\label{sec:conclusion}
In this paper, we present GraphVLM, a comprehensive benchmark for multimodal graph learning (MMGL) that evaluates state-of-the-art methods across six diverse graph datasets. Existing MMGL approaches are grouped into three paradigms based on how Vision-Language Models (VLMs) are utilized: VLM-as-Encoder for GNN-based methods, VLM-as-Aligner for LLM-based methods, and VLM-as-Predictor, where VLMs are used as standalone predictors. Extensive experiments show that fine-tuning VLMs as predictors yields the most significant performance gains, even without explicit graph structural information, highlighting their potential as powerful backbones for MMGL. We hope that GraphVLM serves as a fair and standardized benchmark to facilitate future research and foster continued innovation in this rapidly evolving field.

{
    \small
    \bibliographystyle{ieeenat_fullname}
    \bibliography{main}
}

\newpage
\thispagestyle{empty}
\null
\newpage

\newpage
\section{Appendix}
\label{sec: appendix}

\subsection{Dataset Statistics}
\label{appendix:dataset}
Our experimental benchmark comprises six datasets spanning two distinct domains, with detailed statistical characteristics presented in Table~\ref{table:dataset}.

\begin{table}[htbp]
\centering
\resizebox{0.5\textwidth}{!}{
\begin{tabular}{l|cccccc}
    \toprule
    \textbf{Dataset} & \textbf{\#Nodes} & \textbf{\#Edges} & \textbf{Avg.\#Degree} & \textbf{\#Classes} & \textbf{Domain} \\
    \midrule
    Movies  &16,672&218,390& 26 &19 & E-commerce\\
    Toys &20,695&126,886& 12 &18 & E-commerce \\
    Grocery &  84,379      &   693,154      &     16         &    20   & E-commerce  \\
    Arts  & 28,195 & 197,428 & 14 & 7 & E-commerce\\
    CDs &  36,272       &   844,878      &       47       &     15  & E-commerce  \\
    Reddit &  99,638     &    1,167,188      &    23     &     50  & Social Network  \\
    \bottomrule
\end{tabular}
}
% \caption{Statistics of the datasets.}
\caption{Dataset statistics.}
\label{table:dataset}
\end{table}

\subsection{Baseline Implementations}
\label{appendix:implementation}

We evaluate a diverse set of state-of-the-art methods spanning three major categories of graph learning paradigms.

\begin{itemize}
\item 
\textbf{GNN-based methods.} We include the conventional GCN~\cite{kipf2016gcn}, the widely used GraphSAGE~\cite{hamilton2017graphSAGE}, and a non-graph baseline MLP~\cite{rosenblatt1958perceptron}. For multimodal graph learning, we evaluate MMGCN~\cite{mmgcn} and MGAT~\cite{tao2020mgat}, which explicitly fuse heterogeneous modalities within graph structures. To ensure fair comparison, all models are trained and evaluated under consistent experimental settings, following the MM-Bench framework~\cite{zhu2024multimodalgraphbenchmark}. We additionally adopt the state-of-the-art UniGraph2~\cite{uniGraph2}. However, due to the $O(n^3)$ time complexity of shortest-path distance (SPD) computation in the original implementation, we omit the SPD module for efficiency in our evaluation.

For the CLIP-F-S configuration, we randomly sample five hop-1 neighbors for each anchor node and employ CLIP as the visual encoder within a contrastive learning objective. The training hyperparameters are summarized in Table~\ref{tab:clipfs_param}.

\begin{table}[h]
\centering
\resizebox{0.5\textwidth}{!}{
\begin{tabular}{lcccccc}
\toprule
\textbf{Learning Rate} & \textbf{\#Neighbors} & \textbf{Epoch} & \textbf{Batch Size} & \textbf{Temperature} & \textbf{Optimizer} \\
\midrule
$1 \times 10^{-5}$ & 5 & 1 & 16 & 0.5 & Adam \\
\bottomrule
\end{tabular}
}
\caption{Training hyperparameters for the CLIP-F-S model.}
\label{tab:clipfs_param}
\end{table}

\item 
\textbf{LLM-based methods.} We assess multiple GraphLLMs with distinct architectures, including GraphPrompter~\cite{liu2024graphPrompter}, LLaGA~\cite{chen2024llaga}, GraphGPT~\cite{tang2024graphgpt}, GraphTranslator~\cite{zhang2024graphtranslator}, and the recent MLaGA~\cite{mlaga2025}. For all approaches requiring multimodal embeddings, we employ the pre-trained CLIP~\cite{clip} encoder to extract unified visual–textual representations. All models are evaluated following their official implementations to ensure methodological fidelity.

\item 
\textbf{VLM-based methods.} We further include the state-of-the-art multimodal large language models Qwen-VL-7B~\cite{bai2023qwenvlversatilevisionlanguagemodel}, Qwen2.5-VL-7B~\cite{bai2025qwen25vltechnicalreport} and LLaVA-1.5-7B~\cite{liu2023llava} for predictor experiments, with Qwen-VL-7B also serving as the aligner in RQ2. All models are evaluated under their official zero-shot and fine-tuning protocols. For structural information injection, we select the top-3 most similar nodes based on cosine similarity from the anchor node’s hop-1 neighbors.
\end{itemize}

\subsection{Prompt Templates}
\label{appendix:prompt}
\noindent\textbf{Image summary prompt.} We use Qwen-VL-7B~\cite{bai2023qwenvlversatilevisionlanguagemodel} to generate the image summaries as shown in Table~\ref{table:augmentor_image_prompts}.

\noindent\textbf{Non-structure-aware aligner prompt.}
We synthesize the original textual information with generated image summaries using an LLM, following the prompts in Table~\ref{table :non_structure_aware_modality_synthesis}.

\noindent\textbf{Structure-aware aligner prompt.}
To incorporate structure-aware multimodal information, we design our prompts for modality synthesis as shown in Table~\ref{table: modality_synthesis_structure_aware}.

\noindent\textbf{VLM-as-Predictor prompt design.}
To directly enable VLM as the predictor, we design the prompt accordingly, as shown in Table~\ref{table: VLM_as_predictor_prompt} and Table~\ref{table: VLM_as_predictor_prompt2}. In fine-tuning strategies, the prompt includes \textit{'Assistant: <true label>'}, while in in-context learning, it is excluded.

\subsection{Trade-off between Efficiency and Effectiveness}
We analyze this trade-off from the perspective of cross-domain transferability, with results reported in Table~\ref{tab:transfer}.
VLM-based models achieve substantially higher transfer accuracy than GNN baselines when evaluated on target datasets with or without supervised fine-tuning.
This transfer advantage can help explain the growing interest in LLM/VLM-based graph methods, despite their higher computational cost. We collect the training and inference time on a specific dataset for reference in Table~\ref{tab:efficiency}.

%\begin{table}[t]
\begin{table}[htbp]
\centering
\scriptsize
\renewcommand{\arraystretch}{1.2}   % ← 提高行距
\setlength{\tabcolsep}{1.3pt}         % ← 控制列间距（可选）
\resizebox{1\linewidth}{!}{
\begin{tabular}{l l c c c c}
\toprule
\textbf{Model} & \textbf{Settings} & \textbf{Movies} & \textbf{CDs} & \textbf{Grocery} & \textbf{Arts} \\
\hline
MLP  & Text+Image (CLIP)      & 4.90  & 6.85  & 9.17  & 14.14 \\
GCN  & Text+Image (CLIP)      & 7.69  & 10.05 & 22.02 & 12.55 \\
GraphSAGE & Text+Image (CLIP)      & 3.73  & 6.20  & 5.56  & 20.69 \\
MLP  & Text+Image (CLIP-F-S)  & 5.50  & 8.79  & 14.12 & 18.14 \\
GCN  & Text+Image (CLIP-F-S)  & 3.16  & 7.61  & 8.66  & 18.12 \\
GraphSAGE & Text+Image (CLIP-F-S)  & 1.20  & 9.85  & 11.40 & 18.14 \\
\hline
Qwen2.5-VL & No SFT (Zero-shot)          & \textbf{14.96} & \textbf{34.94} & \textbf{68.79} & \textbf{69.34} \\
Qwen2.5-VL & Text+Image (Non-structure) & \textbf{13.16} & \textbf{37.02} & \textbf{71.95} & \textbf{68.10} \\
Qwen2.5-VL & Text+Image (Structure)    & \textbf{12.17} & \textbf{37.17} & \textbf{71.20} & \textbf{72.26} \\
\bottomrule
\end{tabular}
}
\caption{Transfer ability comparison: Node classification accuracy (\%) when trained on Toys and evaluated on other datasets.}
\label{tab:transfer}
\end{table}

\begin{table}[htbp]
\centering
\scriptsize
\renewcommand{\arraystretch}{1.3}   % ← 提高行距
\setlength{\tabcolsep}{1.6pt}         % ← 控制列间距（可选）
\resizebox{1\linewidth}{!}{
\begin{tabular}{c|cccc|cc|cc}
\toprule
\multirow{2}{*}{\textbf{Stage}}
& \multicolumn{4}{c|}{\textbf{GNN-based}} 
& \multicolumn{2}{c|}{\textbf{LLM-based}} 
& \multicolumn{2}{c}{\textbf{VLM-based}} \\
\cline{2-9}

& GCN & SAGE & MMGCN & MGAT 
& LLaGA & GraphGPT
& Qwen-VL & Qwen2.5-VL \\

\midrule

Training
& \textasciitilde2min & \textasciitilde2min & \textasciitilde3min & \textasciitilde3min
& \textasciitilde17min & \textasciitilde60min
& \textasciitilde5h & \textasciitilde5h \\

Inference
& \textasciitilde10s & \textasciitilde10s & \textasciitilde15s & \textasciitilde20s
& \textasciitilde10min & \textasciitilde30min
& \textasciitilde20min & \textasciitilde20min \\

\bottomrule
\end{tabular}
}
\caption{Efficiency comparison on the Movies dataset.}
\label{tab:efficiency}
\end{table}

\begin{table*}[htbp]
    \centering
    \resizebox{1\textwidth}{!}{%
        \begin{tabular}{p{1.5cm} p{15.8cm}}
            \toprule
            \textbf{Movies:} & \textcolor{Salmon}{<image input>} Given an image of a \textcolor{SkyBlue}{movie} from the \textcolor{SkyBlue}{Amazon movies dataset }, generate a concise and detailed summary. Focus on describing key visual concepts. Ensure the summary is informative and useful for understanding the product as described in user reviews, without losing critical details or introducing unnecessary information. \\
            \midrule
            \textbf{Toys:} & \textcolor{Salmon}{<image input>} Given an image of a \textcolor{SkyBlue}{toy} from the \textcolor{SkyBlue}{Amazon toys dataset }, generate a concise and detailed summary. Focus on describing key visual concepts. Ensure the summary is informative and useful for understanding the product as described in user reviews, without losing critical details or introducing unnecessary information. \\
            \midrule
            \textbf{Grocery:} & \textcolor{Salmon}{<image input>} Given an image of a \textcolor{SkyBlue}{grocery} from the \textcolor{SkyBlue}{Amazon grocery dataset }, generate a concise and detailed summary. Focus on describing key visual concepts. Ensure the summary is informative and useful for understanding the product as described in user reviews, without losing critical details or introducing unnecessary information. \\
            \midrule
            \textbf{CDs:} & \textcolor{Salmon}{<image input>} Given an image of a \textcolor{SkyBlue}{CD} from the \textcolor{SkyBlue}{Amazon CD dataset }, generate a concise and detailed summary. Focus on describing key visual concepts. Ensure the summary is informative and useful for understanding the product as described in user reviews, without losing critical details or introducing unnecessary information. \\
            \midrule
            \textbf{Arts:} & \textcolor{Salmon}{<image input>} Given an image of an \textcolor{SkyBlue}{artwork} from the \textcolor{SkyBlue}{Amazon Art dataset }, generate a concise and detailed summary. Focus on describing key visual concepts. Ensure the summary is informative and useful for understanding the product as described in user reviews, without losing critical details or introducing unnecessary information. \\
            \midrule
            \textbf{Reddit:} & \textcolor{Salmon}{<image input>} Given an image of a \textcolor{SkyBlue}{post} from the \textcolor{SkyBlue}{Reddit dataset}, generate a concise and detailed summary. Focus on describing key visual concepts. Ensure the summary is informative and useful for understanding the post as described in the caption, without losing critical details or introducing unnecessary information. \\
            \bottomrule
        \end{tabular}
    }
    \caption{Prompts used to generate a text description of the image by VLM}
    \label{table:augmentor_image_prompts}
\end{table*}

\begin{table*}[htbp]
    \centering
    \resizebox{1\textwidth}{!}{%
        \begin{tabular}{p{1.5cm} p{15.5cm}}
            \toprule
            \textbf{Movies:} &Given the text information of a product from the \textcolor{SkyBlue}{Amazon Movies} dataset: \textcolor{Salmon}{<text information>}. Image summary: \textcolor{Salmon}{<image summary>} Questions: Using the title, description, and image summary of the product provided above, create an informative and concise description that effectively highlights the product's key features. \\
            \midrule
            \textbf{Toys:} & Given the text information of a product from the \textcolor{SkyBlue}{Amazon toys} dataset: \textcolor{Salmon}{<text information>}. Image summary: \textcolor{Salmon}{<image summary>} Questions: Using the title, description, and image summary of the product provided above, create an informative and concise description that effectively highlights the product's key features. \\ 
            \midrule
            \textbf{Grocery:} & Given the text information of a product from the \textcolor{SkyBlue}{Amazon grocery} dataset: \textcolor{Salmon}{<text information>}. Image summary: \textcolor{Salmon}{<image summary>} Questions: Using the title, description, and image summary of the product provided above, create an informative and concise description that effectively highlights the product's key features. \\ 
            \midrule
            \textbf{CDs:} & Given the text information of a product from the \textcolor{SkyBlue}{Amazon CD} dataset: \textcolor{Salmon}{<text information>}. Image summary: \textcolor{Salmon}{<image summary>} Questions: Using the title, description, and image summary of the product provided above, create an informative and concise description that effectively highlights the product's key features.\\
            \midrule
            \textbf{Arts:} & Given the text information of a product from the \textcolor{SkyBlue}{Amazon Art} dataset: \textcolor{Salmon}{<text information>}. Image summary: \textcolor{Salmon}{<image summary>} Questions: Using the title, description, and image summary of the product provided above, create an informative and concise description that effectively highlights the product's key features.\\
            \midrule
            \textbf{Reddit:} & Given the text information of a post from the \textcolor{SkyBlue}{Reddit} dataset: \textcolor{Salmon}{<text information>}. Image summary: \textcolor{Salmon}{<image summary>} Questions: Using the caption and image summary of the post provided above, create an informative and concise description that effectively highlights the post's key features.\\
            \bottomrule
        \end{tabular}
    }
    
    \caption{Prompts for the non-structure-aware aligner cases.}
    \label{table :non_structure_aware_modality_synthesis}
\end{table*}

\begin{table*}[htbp]
    \centering
    
       \resizebox{1\textwidth}{!}{%
            \begin{tabular}{p{1.5cm} p{17.8cm}}
            \toprule
            \textbf{Movies:} &Given the text information of a product from the \textcolor{SkyBlue}{Amazon movies} dataset: \textcolor{Salmon}{<text information>}. Image summary: \textcolor{Salmon}{<image summary>}. Also given the information of co-purchased or co-reviewed products: text information: \textcolor{Salmon}{<neighbor text information>}, image summary: \textcolor{Salmon}{<neighbor image summary>} (or, if unavailable: 'No co-purchased or co-reviewed product information is available.') Questions: Using the product's title, description, and image summary provided above, along with any co-purchase or co-review data, generate a concise yet informative description of the product.  \\
            \midrule
            \textbf{Toys:} & Given the text information of a product from the \textcolor{SkyBlue}{Amazon toys} dataset: \textcolor{Salmon}{<text information>}. Image summary: \textcolor{Salmon}{<image summary>}. Also given the information of co-purchased or co-reviewed products: text information: \textcolor{Salmon}{<neighbor text information>}, image summary: \textcolor{Salmon}{<neighbor image summary>} (or, if unavailable: 'No co-purchased or co-reviewed product information is available.') Questions: Using the product's title, description, and image summary provided above, along with any co-purchase or co-review data, generate a concise yet informative description of the product. \\
            \midrule
            \textbf{Grocery:} & Given the text information of a product from the \textcolor{SkyBlue}{Amazon grocery} dataset: \textcolor{Salmon}{<text information>}. Image summary: \textcolor{Salmon}{<image summary>}. Also given the information of co-purchased or co-reviewed products: text information: \textcolor{Salmon}{<neighbor text information>}, image summary: \textcolor{Salmon}{<neighbor image summary>} (or, if unavailable: 'No co-purchased or co-reviewed product information is available.') Questions: Using the product's title, description, and image summary provided above, along with any co-purchase or co-review data, generate a concise yet informative description of the product. \\
            \midrule
            \textbf{CDs:} & Given the text information of a product from the \textcolor{SkyBlue}{Amazon CD} dataset: \textcolor{Salmon}{<text information>}. Image summary: \textcolor{Salmon}{<image summary>}. Also given the information of co-purchased or co-reviewed products: text information: \textcolor{Salmon}{<neighbor text information>}, image summary: \textcolor{Salmon}{<neighbor image summary>} (or, if unavailable: 'No co-purchased or co-reviewed product information is available.') Questions: Using the product's title, description, and image summary provided above, along with any co-purchase or co-review data, generate a concise yet informative description of the product.\\
            \midrule
            \textbf{Arts:} & Given the text information of a product from the \textcolor{SkyBlue}{Amazon Art} dataset: \textcolor{Salmon}{<text information>}. Image summary: \textcolor{Salmon}{<image summary>}. Also given the information of co-purchased or co-reviewed products: text information: \textcolor{Salmon}{<neighbor text information>}, image summary: \textcolor{Salmon}{<neighbor image summary>} (or, if unavailable: 'No co-purchased or co-reviewed product information is available.') Questions: Using the product's title, description, and image summary provided above, along with any co-purchase or co-review data, generate a concise yet informative description of the product.\\
            \midrule
            \textbf{Reddit:} & Given the text information of a post from the \textcolor{SkyBlue}{Reddit} dataset: \textcolor{Salmon}{<text information>}. Image summary: \textcolor{Salmon}{<image summary>}. Also given the information of co-commented posts: text information: \textcolor{Salmon}{<neighbor text information>}, image summary: \textcolor{Salmon}{<neighbor image summary>} (or, if unavailable: 'No co-commented post information is available.') Questions: Using the post's caption and image summary provided above, along with any co-commented data, generate a concise yet informative description of the post.\\
            \bottomrule
        \end{tabular}
    }
    \caption{Prompts for the structure-aware aligner cases.}
    \label{table: modality_synthesis_structure_aware}
\end{table*}

\begin{table*}[htbp]
    \centering
    \resizebox{1\textwidth}{!}{%
        \begin{tabular}{p{2.5cm} p{18cm}}
            \toprule
            \textbf{Movies / Toys / Grocery / CDs / Arts:} &Given the target product information on \textcolor{SkyBlue}{Amazon}:
            Picture: \textcolor{Salmon}{<image input>}
            Title and description: \textcolor{Salmon}{<text information>}. 
            Question: Based on the target product's picture, title, and description, which category does the target product belong to? Choose from the following options: \textcolor{Salmon}{<candidates set>}.
            
            Assistant: \textcolor{Salmon}{<truth label>}
            \\
            
            \midrule
                    \textbf{Reddit:} &Given the target post information on \textcolor{SkyBlue}{Reddit}:
            Picture: \textcolor{Salmon}{<image input>}
            Caption: \textcolor{Salmon}{<text information>}.
            Question: Based on the target post's picture and caption, which category does the target post belong to? Choose from the following options: \textcolor{Salmon}{<candidates set>}.
            
            Assistant: \textcolor{Salmon}{<truth label>} \\
            \bottomrule
        \end{tabular}
    }
    \caption{Prompts for the non-structure-aware predictor case.}
    \label{table: VLM_as_predictor_prompt}
\end{table*}

\begin{table*}[htbp]
    \centering
    
    \resizebox{1\textwidth}{!}{%
        \begin{tabular}{p{2.5cm} p{18.5cm}}
            \toprule
            \textbf{Movies / Toys / Grocery / CDs / Arts:} &Given the target product information on \textcolor{SkyBlue}{Amazon}:
            Picture: \textcolor{Salmon}{<image input>}
            Title and description: \textcolor{Salmon}{<text information>}.
            Co-purchased or co-reviewed products: Picture1: \textcolor{Salmon}{<image input>}; Title1: \textcolor{Salmon}{<text information>} ; Picture2: \textcolor{Salmon}{<image input>}; Title2: \textcolor{Salmon}{<text information>} ; Picture3: \textcolor{Salmon}{<image input>}; Title3: \textcolor{Salmon}{<text information>}. 
            Question: Based on the target product's picture, title, description, and related products, which category does the target product belong to? Choose from the following options: \textcolor{Salmon}{<candidates set>}.
            
            Assistant: \textcolor{Salmon}{<truth label>}
            \\
            
            \midrule
                    \textbf{Reddit:} &Given the target post information on \textcolor{SkyBlue}{Reddit}:
            Picture: \textcolor{Salmon}{<image input>}
            Caption: \textcolor{Salmon}{<text information>}.
            Co-commented posts: Picture1: \textcolor{Salmon}{<image input>}; Caption1: \textcolor{Salmon}{<text information>} ; Picture2: \textcolor{Salmon}{<image input>}; Caption2: \textcolor{Salmon}{<text information>} ; Picture3: \textcolor{Salmon}{<image input>}; Caption3: \textcolor{Salmon}{<text information>}.
            Question: Based on the target post's picture, caption, and related posts, which category does the target post belong to? Choose from the following options: \textcolor{Salmon}{<candidates set>}.
            
            Assistant: \textcolor{Salmon}{<truth label>} \\
            \bottomrule
        \end{tabular}
    }
    \caption{Prompts for the structure-aware predictor case.}
    \label{table: VLM_as_predictor_prompt2}
\end{table*}


\end{document}